\documentclass[10pt,journal]{IEEEtran}

\usepackage{booktabs} 

\usepackage{xcolor}
\usepackage{ifthen}
\usepackage{microtype}
\PassOptionsToPackage{warn}{textcomp}
\usepackage{textcomp} 
\usepackage{url}

\usepackage{cite}
\usepackage{tabu}
\usepackage{booktabs}
\usepackage{amsmath}
\usepackage{amssymb}
\usepackage{subfigure}
\usepackage{graphicx}
\usepackage{soul}
\usepackage{algorithmic}
\usepackage{algorithm}
\usepackage{ragged2e}
\usepackage{xspace}
\usepackage{bbding}

\usepackage{color}
\definecolor{quill}{rgb}{0.69,0.61,0.85}
\definecolor{contentchange}{rgb}{0.9,0.61,0.5}
\newcommand{\shortname}{ArtCrafter\xspace}
\newcommand{\datasetname}{ArtMarket\xspace}
\newcommand{\etal}{{\sl et al.}}
 
\begin{document}

\title{ArtCrafter: Text-Image Aligning Style Transfer via Embedding Reframing}

\author{Nisha Huang,
        Kaer Huang,
        Yifan Pu,
        Jiangshan Wang,
        Jie Guo,
        Yiqiang Yan, \\
        Xiu Li,~\IEEEmembership{Member,~IEEE},
        Tong-Yee Lee,~\IEEEmembership{Senior Member,~IEEE}

\IEEEcompsocitemizethanks{
\IEEEcompsocthanksitem  N. Huang, Y. Pu, J. Wang, and X. Li are with Tsinghua International Graduate School, Tsinghua University, Shenzhen 518071, China. E-mail: \{hns24, puyf23, wjs23\}@mails.tsinghua.edu.cn and li.xiu@sz.tsinghua.edu.cn.
\IEEEcompsocthanksitem  N. Huang and G. Jie are also with Pengcheng Laboratory, Shenzhen 518055, China. E-mail: \{huangnsh, guoj01\}@pcl.ac.cn
\IEEEcompsocthanksitem K. Huang and Y. Yan are with Lenovo, Inc. E-mail: \{huangke1, yanyq\}@lenovo.com.
\IEEEcompsocthanksitem T.-Y. Lee is with National Cheng Kung University, Tainan 701, Taiwan.\protect\\
E-mail: tonylee@mail.ncku.edu.tw.}
\thanks{This work was supported by the National Science and Technology Council under no. 113-2221-E-006-161-MY3, Taiwan.
(Corresponding author: Xiu Li.)}
}



\maketitle


\begin{abstract}
\justifying
Recent years have witnessed significant advancements in text-guided style transfer, primarily attributed to innovations in diffusion models. 
These models excel in conditional guidance, utilizing text or images to direct the sampling process.
However, despite their capabilities, direct conditional guidance approaches often face challenges in balancing the expressiveness of textual semantics with the diversity of output results while capturing stylistic features.
To address these challenges, we introduce \shortname, a novel framework for text-to-image style transfer.
Specifically, we introduce an attention-based style extraction module, meticulously engineered to capture the subtle stylistic elements within an image. This module features a multi-layer architecture that leverages the capabilities of perceiver attention mechanisms to integrate fine-grained information.
Additionally, we present a novel text-image aligning augmentation component that adeptly balances control over both modalities, enabling the model to efficiently map image and text embeddings into a shared feature space. 
We achieve this through attention operations that enable smooth information flow between modalities.
Lastly, we incorporate an explicit modulation that seamlessly blends multimodal enhanced embeddings with original embeddings through an embedding reframing design, empowering the model to generate diverse outputs.
Extensive experiments demonstrate that \shortname yields impressive results in visual stylization, exhibiting exceptional levels of stylistic intensity, controllability, and diversity.
\end{abstract}
\begin{figure*}
\centering
\includegraphics[width=\linewidth]{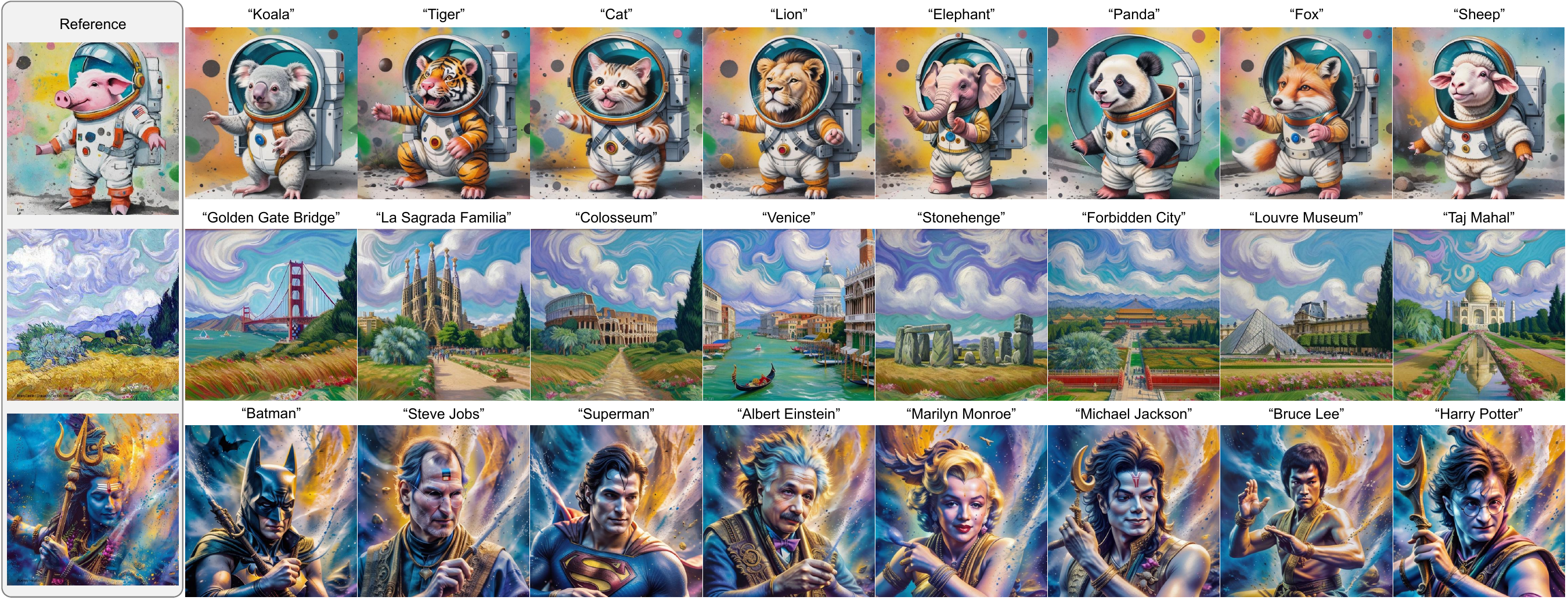}
\caption{\textbf{ArtCrafter generation results.} By injecting the features of the reference images and text prompts during the diffusion process, our method is capable of capturing and generating a faithful style representation. 
}
\label{fig:teaser}
\end{figure*}

\section{Introduction}
\label{sec:intro}
\IEEEPARstart{D}iffusion-based text-to-image generation models~\cite{latentdiffusion,sdxl} have made significant strides in the areas of personalization and customization, particularly in consistent synthesis tasks such as identity protection~\cite{photomaker,instantid}, object customization~\cite{RealCustom,ma2024subject}, and style transfer~\cite{styleid,inst,artadapter,z-star,zhang2024artbank,jiang2023avatarcraft,huang2024creativesynth}. Among these applications, text-guided style transfer has emerged as a powerful tool, focusing on fine-grained style representation that captures abstract concepts like texture, color, composition, brushstroke, and genre. This approach enables the creation of a diverse range of personalized outputs that are deeply rooted in the semantic essence of the input text.

\begin{figure}
\centering
\includegraphics[width= \linewidth]{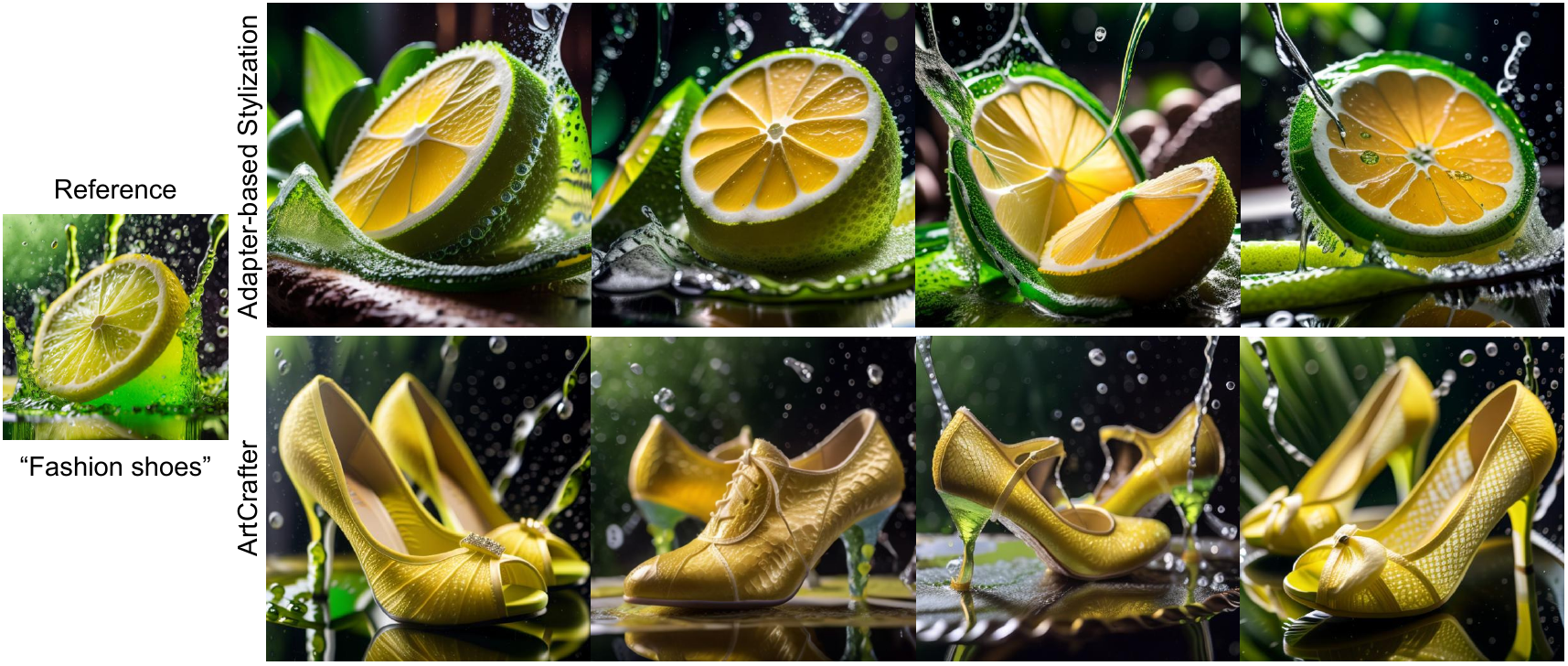}
\caption{\textbf{Generic adapter-based vs. \shortname generation results.}  Given a content description of ``Fashion Shoes'', generic adapter-based generation (above) results in unaligned results and limited result diversity. In contrast, our approach (below) generates text-aligned content as well as multiple shoe types.}
\label{fig:insight}
\end{figure}

Current methods for stylization tasks often leverage pre-trained diffusion models~\cite{liu2023stylecrafter,faceadapter,ipadapter,styleshot,instantstyle,Ctrl-X,ipadapter-Instruct,wang2024gra,huang2024diffstyler}, enhancing model features by adding a trainable adapter module without full retraining. In text-to-image style transfer applications, adapter-based methods shape the style and content of the output by adjusting the condition guidance scales over the input image and text prompts. However, we have identified three main issues with previous research:
\textbf{1)} \textit{Inadequate artistic style representation.} Traditional image encoder architectures and training data, which primarily focus on natural images, limit their ability to capture the intricate textures and stylistic nuances of artistic images.
\textbf{2)} \textit{Suboptimal text-guided conditions.} As shown in Fig.~\ref{fig:insight}, the usual adapter-based method~\cite{ipadapter} fails to deliver the expected results when using the text condition \textit{``Fashion shoes''}. This discrepancy arises because the amount of information in image and text embeddings is not equivalent, yet adapter-based methods~\cite{liu2023stylecrafter,faceadapter,ipadapter,styleshot,instantstyle,Ctrl-X,ipadapter-Instruct} often directly concatenate the two without addressing the imbalance and disparity, leading to image data overshadowing text prompts during the sampling process.
\textbf{3)} \textit{Lack of output diversity.} The constrained liberation of textual guidance results in generated outputs that closely resemble the reference images, thereby limiting the diversity of the results.

To address the aforementioned challenges, we introduce \shortname, a novel embedding reframing solution based on diffusion models, specifically tailored for text-guided stylization tasks. \shortname comprises three key components:
\textbf{1)} \textit{Attention-based Style Extraction} (Sec.~\ref{sec:3.2}): This component leverages multi-level features to capture intricate stylistic details, ensuring more coherent and accurate stylistic encoding. It employs a non-layer refinement module and a multi-layer attention architecture to capture both local and global stylistic elements. Additionally, we introduce the \datasetname dataset, which pairs art images with descriptive texts, enabling us to fine-tune an encoder initially trained on natural images. This approach retains strong generalization capabilities while being sensitive to the unique visual elements of artistic styles.
\textbf{2)} \textit{Text-Image Aligning Augmentation} (Sec.~\ref{sec:3.3}): This module enhances the alignment of image and text embeddings through crafted attentional interactions, mapping them into a shared feature space. This alignment ensures that the generated images reflect both the style of the reference image and the content of the textual conditions, thereby enhancing controllability.
\textbf{3)} \textit{Explicit Modulation} (Sec.~\ref{sec:3.4}): This component enhances the adaptability of conventional fusion techniques through the implementation of linear interpolation and concatenation schemes. This method facilitates the merging of original image and text embeddings with multimodal embeddings, yielding enhanced embeddings. It allows \shortname to generate images that are relevant to the text prompts and exhibit diverse visual representations.
Our main contributions are summarized as follows:
\begin{itemize}
    \item[--] We introduce \shortname, a lightweight adapter designed to enhance the capabilities of pre-trained diffusion models in text-guided stylized image generation. Our approach focuses on attention-based style feature extraction, effectively capturing both local and global features.
    \item[--] We propose an innovative text-image aligning augmentation module that enables robust interaction between reference images and textual descriptions within a shared feature space, significantly enhancing the influence of text prompts on the generative process.
    \item[--] The explicit modulation within \shortname optimizes the utilization of multimodal embeddings, offering greater flexibility and diversity than conventional methods. Additionally, \shortname is compatible with additional control conditions and achieves superior performance across various experimental benchmarks compared to state-of-the-art approaches.
\end{itemize}

\begin{figure*}
\centering
\includegraphics[width=\linewidth]{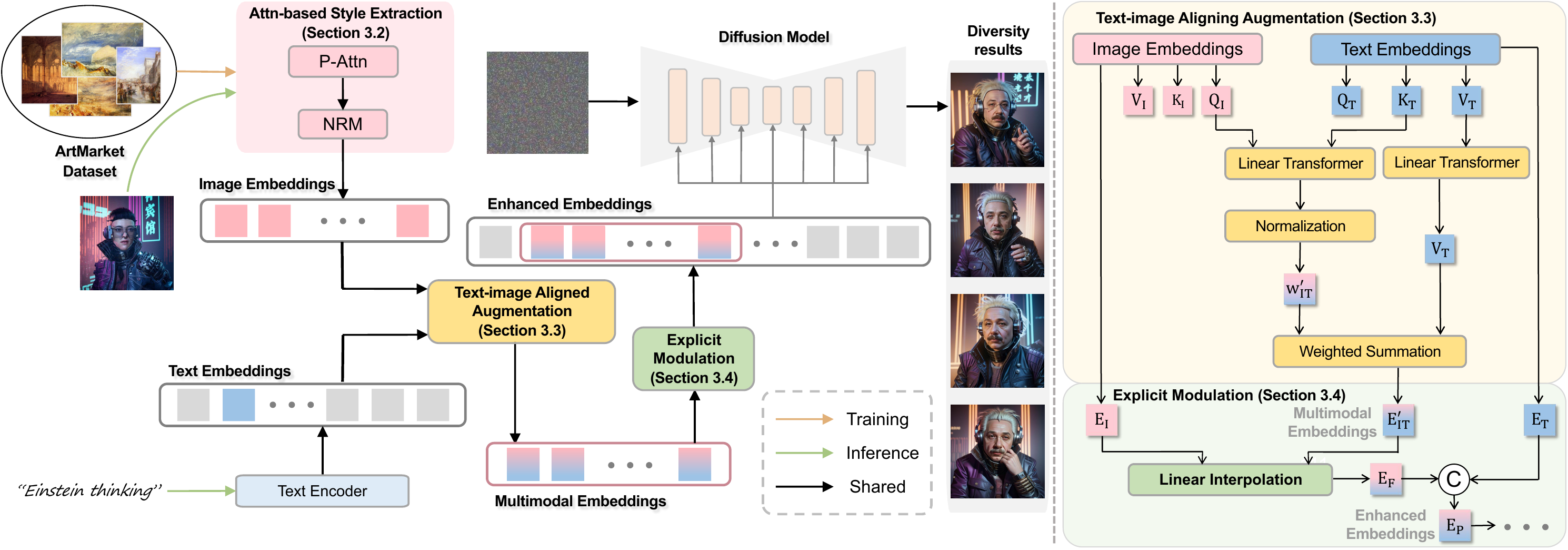}
\caption{\textbf{Our proposed \shortname's overview pipeline consists of three modules}: 1) Attention-based style extraction captures fine-grained style features from images using multi-head perceiver attention and feed-forward networks. 2) Text-image aligning augmentation transfers the style embedding to the textual space to improve the image-text consistency in the generated images. 3) Explicit modulation seamlessly combines multimodal embeddings with original image embeddings and text embeddings in different ways, increasing the versatility and diversity of the method.
}
\label{fig:pipeline}
\end{figure*}

\section{Related Work}
\label{sec:relatedwork}
\subsection{Attention Control in diffusion models.}
Following the remarkable progress made in the field of pre-training text-to-image diffusion models~\cite{sdxl,latentdiffusion,singh2023high,ddpm}, a series of image editing efforts~\cite{controlnet,instructpix2pix,csgo,blendeddiffusion} have emerged.
Hertz \etal~\cite{Prompt-to-Prompt} propose the Prompt-to-Prompt method, which achieves text-based partial image editing and generates edited images that conform to textual conditions by replacing original vocabulary and cross-attention maps.
Plug-and-play~\cite{Plug-and-play} utilizes the spatial features and self-attention mapping of the original image to guide the diffusion model for text-guided image-to-image translation while preserving the spatial layout of the original image.
Later, MasaCtrl~\cite{masactrl} proposes a mutual self-attention control technique for coherent image editing. Consistent image editing is achieved by preserving the key and value of the self-attention layer of the source image while conditioning the model with desired text prompts.
Recently, StyleID~\cite{styleid} proposes a style migration method without training by injecting styles in the self-attention layer and introduces query preservation, attention temperature scaling, and initial latent AdaIN techniques to minimize the impact of style injection on the original content.
$\mathcal{Z}^*$~\cite{z-star} shows how to extract style information directly from style images and fuse it with content images without training by means of an attentional reweighting strategy.
Unlike the methods mentioned above, our approach focuses on the interaction of image and text information, as well as a balance for guiding the generation process.

\subsection{Text-guided style transfer.}
Stylization~\cite{deadiff,liu2023name,Ganugula_2023_ICCV,yang20233dstyle} is to control the content through text and make the generated image consistent while keeping the style of the reference image.
StyleDrop~\cite{styledrop} is a method of achieving arbitrary style synthesis from a small number of stylized images and textual descriptions by efficiently fine-tuning a small number of parameters of a pre-trained model and combining iterative training with feedback to improve the quality of the generated images.
The well-received work IP-Adapter~\cite{ipadapter} improves stylization image generation by introducing the embedding of input images in an additional layer of cross-attention, which enhances the model's ability to capture features from the input images.
Building on the IP-Adapter, InstantStyle~\cite{instantstyle} manually selects specific attention layers to control the style of the output. However, for the IP-Adapter conditioned on natural images, the expected conditioning of the input artistic image does not always work.
Visual Style Prompting (VSP)~\cite{vsp} captures stylistic details by fusing key features of the reference image in a later self-attention layer while preserving the content consistency of the original image. However, compared to cross-attention, self-attention provides a weaker ability to control the semantic content of the generated image.
Style Aligned~\cite{stylealigned} attempts to align style and content through a shared attention mechanism. However, it generates results with content information leaked from the style image, and there are challenges in disentangling content and style.
StyleShot~\cite{styleshot} is trained by a two-stage style control method. However, detailed information within the control is easily lost due to sparse rows and columns. 
Furthermore, \cite{styleshot}, \cite{liu2023stylecrafter}, and \cite{csgo} only trained on the art-text datasets, making it difficult to broadly adapt to arbitrary stylistic features.

\section{Method}
The overall architecture of \shortname is shown in Fig.~\ref{fig:pipeline}.
As reviewed in Sec.~\ref{sec:3.1}, \shortname is built upon diffusion model~\cite{latentdiffusion,realistic}. In Section~\ref{sec:3.2}, we introduce attention-based style extraction, which captures multi-level style information by non-layer refinement module and multi-layer design. Text-image aligning augmentation (Section~\ref{sec:3.3}) allows the model to dynamically weigh the importance of different parts of the text prompt. This enables a more nuanced and context-aware generation process, resulting in images that are more closely related to the text prompt. Explicit modulation (Section~\ref{sec:3.4}) effectively combines textual and visual information, enabling the model to generate images that are relevant to the text prompt and have diverse visual representations.
In Section~\ref{sec:3.5}, we provide a detailed description of the training and inference processes. During training, we optimize the adapter while keeping the parameters of the pre-trained diffusion model frozen. The adapter is trained on the ArtMarket dataset. During inference, we enhance controllability through classifier-free guidance with dual conditioning, which enables the model to generate images more aligned with the text prompt while retaining the stylistic characteristics of the art image.

\subsection{Preliminary}
\label{sec:3.1}


The forward diffusion process incrementally adds Gaussian noise \( \epsilon \) to the data \( x_0 \) through a Markov chain. Specifically, the data \( x_0 \) is gradually corrupted by noise over \( T \) timesteps, resulting in a sequence of progressively noisier data points \( x_0, x_1, \ldots, x_T \). Each step \( t \) in the Markov chain is defined by:
\begin{equation}
x_t = \sqrt{1 - \beta_t} x_{t-1} + \sqrt{\beta_t} \epsilon,
\end{equation}
where \( \beta_t \) is a predefined noise schedule that controls the amount of noise added at each step, and \( \epsilon \sim \mathcal{N}(0, 1) \) is Gaussian noise. By the final step \( T \), the data \( x_T \) is typically indistinguishable from pure Gaussian noise.

The reverse denoising process aims to reconstruct the original data from the noisy data \( x_T \). This process is driven by a learnable denoising model \( \epsilon_\theta(x_t, t, c) \) parameterized by \( \theta \). The denoising model is implemented using a U-Net architecture~\cite{unet}, which is capable of capturing complex patterns and dependencies in the data. The denoising process starts from \( x_T \sim \mathcal{N}(0, 1) \) and iteratively refines the data by predicting and removing the noise at each timestep \( t \):
\begin{equation}
\hat{x}_{t-1} = \sqrt{1 - \beta_t} \hat{x}_t - \sqrt{\beta_t} \epsilon_\theta(\hat{x}_t, t, c).
\end{equation}
The denoising model \( \epsilon_\theta(\cdot) \) is trained to predict the noise \( \epsilon \) added at each step, allowing the model to progressively reconstruct the original data \( x_0 \).

The denoising model \( \epsilon_\theta(\cdot) \) is trained with a mean-squared loss derived from a simplified variant of the variational bound:
\begin{equation}
\mathcal{L} = \mathbb{E}_{t, \mathbf{x}_0, \epsilon}\left[ \|\epsilon - \hat{\epsilon}_\theta(\mathbf{x}_t, t, c)\|^2 \right],
\end{equation}
where \( c \) denotes an optional condition. In the diffusion model, \( c \) is generally represented by the text embeddings \( E_T \) encoded from a text prompt using CLIP~\cite{clip}. These text embeddings are integrated into the diffusion model, allowing the model to generate samples conditioned on the provided text prompt.
This conditioning mechanism enables the diffusion model to produce outputs that are semantically aligned with the text description, making it a powerful tool for text-to-image synthesis and other conditional generation tasks.

\begin{figure*}
\centering
\includegraphics[width= \linewidth]{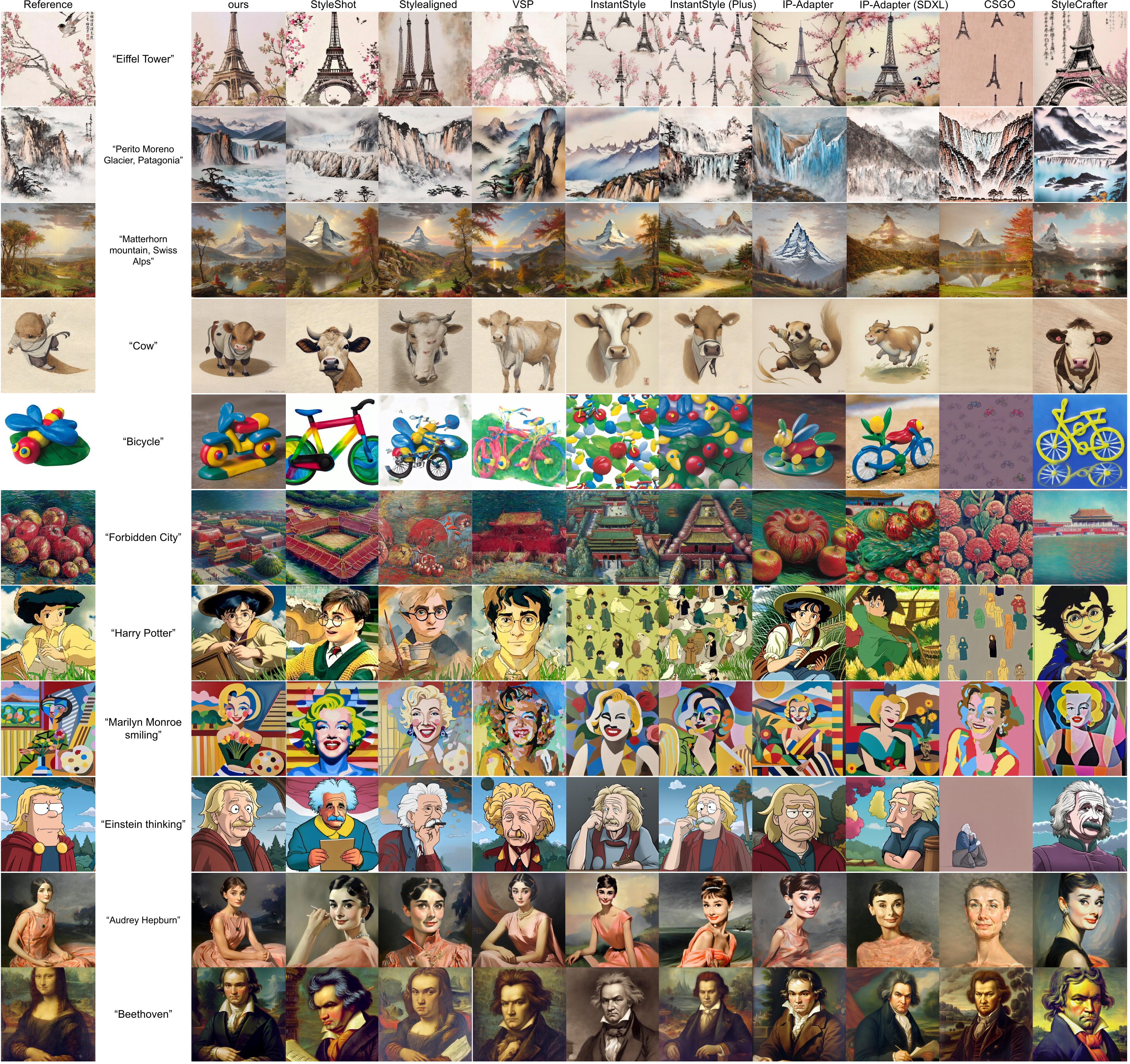}
\caption{\textbf{Qualitative Comparison with Other State-of-the-Art Text-Guided Stylization Methods.} We conducted a comprehensive qualitative evaluation by comparing our method with various state-of-the-art text-guided stylization methods, including Styleshot~\cite{styleshot}, Style Aligned~\cite{stylealigned}, VSP~\cite{vsp}, InstantStyle~\cite{instantstyle}, InstantStyle (Plus)~\cite{instantstyle}, IP-Adapter~\cite{ipadapter}, IP-Adapter (SDXL)~\cite{ipadapter}, CSGO~\cite{csgo}, and StyleCrafter~\cite{liu2023stylecrafter}.
}
\label{fig:compare}
\end{figure*}


\subsection{Attention-based Style Extraction (ASE)}
\label{sec:3.2}
ASE enhances the style encoding capabilities by integrating fine-grained features through a multi-layer architecture. Unlike the methods in \cite{perceiver,ffn}, which focus on general feature extraction, our ASE method specifically targets intricate style details from images by leveraging Perceiver Attention~\cite{perceiver} (P-Attn) and a Non-Linear Refinement Module (NRM). Our technical contributions lie in the novel combination and adaptation of these components to style extraction, which significantly enhances the representation of stylistic characteristics. The method involves several key steps: initializing latent variables, expanding them to match the input batch size, applying P-Attn, and using the NRM to refine the latent variables.

Given the reference image, we obtain the input image embeddings through CLIP, denoted as \( x \). The latent variables \( z \) are initialized as a tensor with a shape of \( (1, N, D) \), where \( N \) is the number of queries and \( D \) is the dimension of the latent space. To stabilize the training process, the latent variables are normalized by dividing by the square root of \( D \):
\begin{equation}
z = \frac{\mathcal{N}(0, 1)}{D^{0.5}}.
\end{equation}
To match the batch size of the input \( x \), we expand the latent variable \( z \) by repeating it along the batch dimension. This process can be represented as:
\begin{equation}
z = z \otimes \mathbf{1}_B,
\end{equation}
where \( \mathbf{1}_B \) is a tensor of ones with shape \( (B, 1, 1) \), and \( B \) is the batch size of \( x \). This operation results in a tensor \( z \) with shape \( (B, N, D) \), where \( N \) is the number of queries and \( D \) is the dimension of the latent space.

The P-Attn mechanism denoted as \( \text{P-Attn} \), is then applied to update the latent variables by attending to the input \( x \) and the repeated latent variables:
\begin{equation}
\text{P-Attn}(x, z) = \text{softmax}\left(\frac{z x^T}{\sqrt{d_k}}\right) \cdot x,
\end{equation}
\begin{equation}
z' = \text{P-Attn}(x, z) + z,
\end{equation}
where \( d_k \) is the dimension of the key tensor, typically equal to \( D \). This operation allows the model to selectively focus on different parts of the input data based on the learnable latent variables, thereby capturing intricate style details from the reference image. Compared to \cite{perceiver,ffn}, our use of P-Attn in the context of style extraction is novel and specifically tailored to enhance the representation of stylistic nuances.

The Non-Linear Refinement Module (NRM) consists of two linear transformations with a GELU activation function in between:
\begin{equation}
\text{NRM}(z') = W_2 \cdot \text{GELU}(W_1 \cdot z' + b_1) + b_2,
\end{equation}
where \( W_1 \) and \( W_2 \) are weight matrices, and \( b_1 \) and \( b_2 \) are bias terms. The NRM further refines the latent variables by applying non-linear transformations, enhancing the representation of the style features. In contrast to \cite{perceiver,ffn}, our NRM is designed to specifically refine style-related latent variables, resulting in a more detailed and robust style embedding.

The output \( E_I \) represents the style embeddings extracted from the input image, obtained by combining the NRM output and the updated latent variables \( z' \):
\begin{equation}
E_I = \text{NRM}(z') + z'.
\end{equation}
This final step integrates the refined latent variables with the non-linear transformations from the NRM, resulting in a robust and detailed style embedding that captures the intricate style details from the reference image. This style embedding can be used to guide the generation process in downstream applications, ensuring that the generated outputs inherit the desired stylistic characteristics.

\subsection{Text-Image Aligning Augmentation (TIAA)}
\label{sec:3.3}

The TIAA method is designed to dynamically prioritize different aspects of the text prompt by leveraging cross-attention mechanisms. This module allows the model to integrate image and text embeddings more effectively, projecting them into a shared feature space where their interactions can be more nuanced. Our novel process involves three main steps: linear transformation, attention-weight calculation, and attention-weighted value matrix computation.

We start by transforming the image prompt embeddings \( E_{\text{I}} \) and the text prompt embeddings \( E_{\text{T}} \) into query, key, and value matrices through linear layers. These transformations are represented as:
\begin{equation}
Q_{\text{I}} = W_{Q_{\text{I}}} E_{\text{I}}, \quad K_{\text{T}} = W_{K_{\text{T}}} E_{\text{T}}, \quad V_{\text{T}} = W_{V_{\text{T}}} E_{\text{T}}.
\end{equation}
Here, \( W_{Q_{\text{I}}} \), \( W_{K_{\text{T}}} \), and \( W_{V_{\text{T}}} \) are the weight matrices associated with the query for images, key for text, and value for text, respectively. These linear transformations map the embeddings into a shared feature space, enabling the subsequent attention mechanism to effectively capture the interactions between the image and text representations.

The attention weights \( \mathbf{w}_{\text{IT}} \) are calculated by taking the dot product of the query matrix \( Q_{\text{I}} \) and the key matrix \( K_{\text{T}} \), scaled by the square root of the key dimension \( d_{k_{\text{T}}} \) to prevent gradient disappearance:
\begin{equation}
\mathbf{w}_{\text{IT}} = \frac{Q_{\text{I}} K_{\text{T}}^T}{\sqrt{d_{k_{\text{T}}}}}.
\end{equation}
The softmax function is then applied to these raw attention scores to obtain the normalized attention weights \( \mathbf{w'}_{\text{IT}} \):
\begin{equation}
\mathbf{w'}_{\text{IT}} = \text{softmax}(\mathbf{w}_{\text{IT}}).
\end{equation}
The normalized attention weights \( \mathbf{w'}_{\text{IT}} \) represent the importance of each text feature with respect to the image features, allowing the model to dynamically focus on the most relevant parts of the text prompt.

Using the normalized attention weights \( \mathbf{w'}_{\text{IT}} \), we compute the weighted sum of the value matrix \( V_{\text{T}} \) to generate the multimodal embeddings \( E'_{\text{IT}} \):
\begin{equation}
E'_{\text{IT}} = \mathbf{w'}_{\text{IT}} \cdot V_{\text{T}}.
\end{equation}
The resulting multimodal embeddings \( E'_{\text{IT}} \) capture the multimodal context more effectively, allowing the model to generate images that are more closely aligned with the semantic content of the text prompt. From our experiments, this approach is particularly beneficial in scenarios where the textual and visual information needs to be tightly integrated to produce coherent outputs. By leveraging cross-attention mechanisms, the model can dynamically prioritize different aspects of the text prompt, leading to more accurate and contextually relevant image generation.

\begin{table*}
\centering
\footnotesize
\caption{
\textbf{Quantitative Comparison.} To provide a comprehensive evaluation of our method against other state-of-the-art text-guided stylization techniques, we conducted extensive quantitative experiments using a diverse set of metrics. These metrics cover key aspects of image quality, including image consistency (measured by CLIP-Image and DINO-v2), text consistency (assessed via CLIP-Text), and diversity (quantified using LPIPS). The best results are highlighted in \textbf{bold}, and the second-best results are marked with \underline{underline}.
}
\setlength{\tabcolsep}{4pt} 
\begin{tabular}{c|cccccccccc}
\toprule
{\textbf{Metric}} & \textbf{Ours} & \textbf{Styleshot} & \textbf{Style Aligned} & \textbf{VSP} & \textbf{InstantSt.} & \textbf{InstantSt. (Plus)} & \textbf{IP-Ada.} & \textbf{IP-Ada. (SDXL)}  & \textbf{CSGO} & \textbf{StyleCrafter}\\
\midrule
{CLIP-Text $\uparrow$} & \textbf{22.57}   & \underline{22.46} & 18.26 & 22.29 & 19.78 & 19.53 & 15.01 & 19.14 &17.82  &19.37\\
{CLIP-Image $\uparrow$ }   & \textbf{69.48}   & 59.47 & 63.82 & 66.31 & 62.59 & 64.65 & \underline{68.90} & 67.43 &55.16  &58.68\\
{DINO-v2 $\uparrow$ }   & \textbf{40.92}   & 23.45 & 27.89 & 35.34 & 29.76 & 30.62 & \underline{38.12} & 32.98 &24.53 &31.60 \\
{LPIPS $\uparrow$ }   & \textbf{0.4908}   & \underline{0.4561} & 0.3892 & 0.2653 & 0.1478 & 0.1234 & 0.3456 & 0.2987 &0.3655 &0.3194 \\
\bottomrule
\end{tabular}
\label{tab_quan_con}
\end{table*}
\begin{table*}
\centering
\footnotesize
\caption{
\textbf{User Study.} To further validate the effectiveness of our method from a human perception perspective, we conducted a comprehensive user study. Participants were asked to evaluate the generated images based on three key criteria: content consistency (Human-Content), style consistency (Human-Style), and overall quality (Human-Overall). The best results are in \textbf{bold} while the second-best results are marked with \underline{underline}.
}
\setlength{\tabcolsep}{4pt} 
\begin{tabular}{c|cccccccccc}
\toprule
\textbf{Metric} & \textbf{Ours} & \textbf{Styleshot} & \textbf{Style Aligned} & \textbf{VSP} & \textbf{InstantSt.} & \textbf{InstantSt. (Plus)} & \textbf{IP-Ada.} & \textbf{IP-Ada. (SDXL)}  & \textbf{CSGO} & \textbf{StyleCrafter}\\
\midrule
{Human-Content $\uparrow$} & \textbf{4.27}  & \underline{3.96}  &3.19  &3.85  &3.08  &2.72  &2.92  &3.27   &2.38   &3.42 \\
{Human-Style $\uparrow$ }  & \textbf{4.08} & 3.08 & 3.04  &3.58  &2.92  &3.15  &\underline{3.81}  &3.46   &2.65   &2.81  \\
{Human-Overall $\uparrow$ } & \textbf{4.19} & \underline{3.73}  & 3.23  & 3.65  &2.81  &2.96  &3.50  &3.15  &2.54   &2.88 \\
\bottomrule
\end{tabular}
\label{tab_quan_userstudy}
\end{table*}

\subsection{Explicit Modulation (EM)}
\label{sec:3.4}

Traditional stylization methods lack flexibility. For example, the adapter-based method in Fig.~\ref{fig:insight} generates similar results, failing to produce diverse outputs. In contrast, the EM method integrates image embeddings with multimodal embeddings through linear interpolation, flexibly balancing both influences. This significantly enhances the diversity of the output.


Specifically, we fuse the image embeddings \( E_{\text{I}} \) with the multimodal embeddings \( E'_{\text{IT}} \) through linear interpolation:
\begin{equation}
E_{\text{F}} = \alpha E_{\text{I}} + (1 - \alpha) E'_{\text{IT}},
\end{equation}
where \( \alpha \) is a predefined constant controlling the fusion ratio between the original and enhanced embeddings. The value of \( \alpha \) can be adjusted to balance the contribution of the image embeddings and the multimodal embeddings, ensuring that the fused embeddings \( E_{\text{F}} \) retain the essential features of both modalities.

Ultimately, we concatenate the fused image embeddings \( E_{\text{F}} \) with the text prompt embeddings \( E_{\text{T}} \) to form the complete prompt embeddings for image generation:
\begin{equation}
E_{\text{P}} = E_{\text{T}} \oplus E_{\text{F}},
\end{equation}
where \( \oplus \) denotes the concatenation operation. The resulting prompt embeddings \( E_{\text{P}} \) integrate the enhanced multimodal information into the diffusion model. By carefully balancing the contributions of the text and image embeddings, the model gains a robust and controlled representation that effectively captures multimodal conditions, thereby improving the overall generation performance.

\subsection{Training and Inference}
\label{sec:3.5}
During training, we optimize the adapter while keeping the parameters of the pre-trained diffusion model frozen. The adapter is trained on the ArtMarket dataset, using the same loss function as the original Stable Diffusion (SD):
\begin{equation}
\mathcal{L} = \mathbb{E}_{z_0, \epsilon, c_t, c_{\text{art}}, t} \|\epsilon - \epsilon_{\theta}(z_t, c_t, c_{\text{art}}, t)\|^2.
\end{equation}
Here, $\epsilon$ is the randomly sampled Gaussian noise, $\epsilon_{\theta}$ is the noise prediction model, and $t$ is the timestep. Note that during training, the latent variable $z$ is constructed with the art image $c_{\text{art}}$ as follows:
\begin{equation}
z_t = \sqrt{\bar{\alpha}_t} \psi(c_{\text{art}}) + \sqrt{1 - \bar{\alpha}_t} \epsilon,
\end{equation}
where $\psi(\cdot)$ is the function mapping the original input to the latent space, and $\bar{\alpha}_t$ represents the cumulative variance at timestep $t$.

Classifier-free guidance~\cite{ho2021classifier} is a well-established technique in the field of diffusion models.
During the inference phase, we enhance controllability through classifier-free guidance with dual conditioning. The output at timestep $t$ is given by:
\begin{equation}
\hat{\epsilon}_{\theta}(x_t, c_t, c_{\text{art}}, t) = w \epsilon_{\theta}(x_t, c_t, c_{\text{art}}, t) + (1 - w) \epsilon_{\theta}(x_t, t),
\end{equation}
where the classifier-free guidance factor $w$ is set to $0.6$.

\section{Experiments}
\subsection{Experimental Setup}
We train \shortname on a training data consisting of about $500,000$ real image-text pairs from LAION-Aesthetic~\cite{laion} and $50k$ art-text pairs from our proposed \datasetname dataset.
The images are paired with text descriptions generated by BLIP-2~\cite{blip2}, forming image-text data pairs. More explanations and examples of the data are given in the Supplementary Materials.
During both training and inference, we resize the input images to a spatial resolution of $512 \times 512$. 
We have implemented our method over the stable diffusion 1.5 version~\cite{latentdiffusion}.
Our training processes are conducted using $8$ NVIDIA A100 GPUs, each with $80GB$ of memory, and a batch size of $8$ per GPU.
The inference phase, which consumes $5185$ MiB of memory, takes about one second of sampling time on a single $A100$ at denoising steps of $50$.

\subsection{Qualitative Evaluations}
We evaluate our proposed method by comparing it with various existing methods, including but not limited to Styleshot~\cite{styleshot}, Style Aligned~\cite{stylealigned}, VSP~\cite{vsp}, InstantStyle~\cite{instantstyle}, InstantStyle (Plus)~\cite{instantstyle,sdxl}, IP-Adapter~\cite{ipadapter}, IP-Adapter (SDXL)~\cite{ipadapter,sdxl}, CSGO~\cite{csgo}, and StyleCrafter~\cite{liu2023stylecrafter}. We utilized the publicly available implementations of these methods and followed their recommended configurations for testing.

The qualitative comparison presented in Fig.~\ref{fig:compare} offers a visual assessment of the results achieved by various stylization methods.
StyleShot shows some deficiencies in style representation, particularly in the capture of details and the consistency of style.
Style Aligned and VSP sometimes exhibit discrepancies between the output style and the input style reference, which may lead to the loss of stylistic features.
Because the stylistic and semantic information in the style map is not completely separated, InstantStyle and its plus version deliver repeated non-essential semantic information in the results, such as the ``Harry Potter'' and ``Bicycle'' examples. 
The content information in the style image also affects the IP-Adapter and its SDXL version; for instance, the ``apple'' content is introduced in the ``Forbidden City, Beijing'' case. 
The fidelity of the results generated by CSGO is constrained by the number of art datasets. 
StyleCrafter produces outputs that are more consistent with the text, although the styles differ substantially from the reference image.
As illustrated in Fig.~\ref{fig:compare}, our method achieves captivating stylization by effectively transferring style patterns (such as brush strokes and lines). These style patterns are skillfully adapted to the content semantics, as seen in rows 1, 3, and 7 of Fig.~\ref{fig:compare}.

\begin{figure}
\centering
\includegraphics[width= \linewidth]{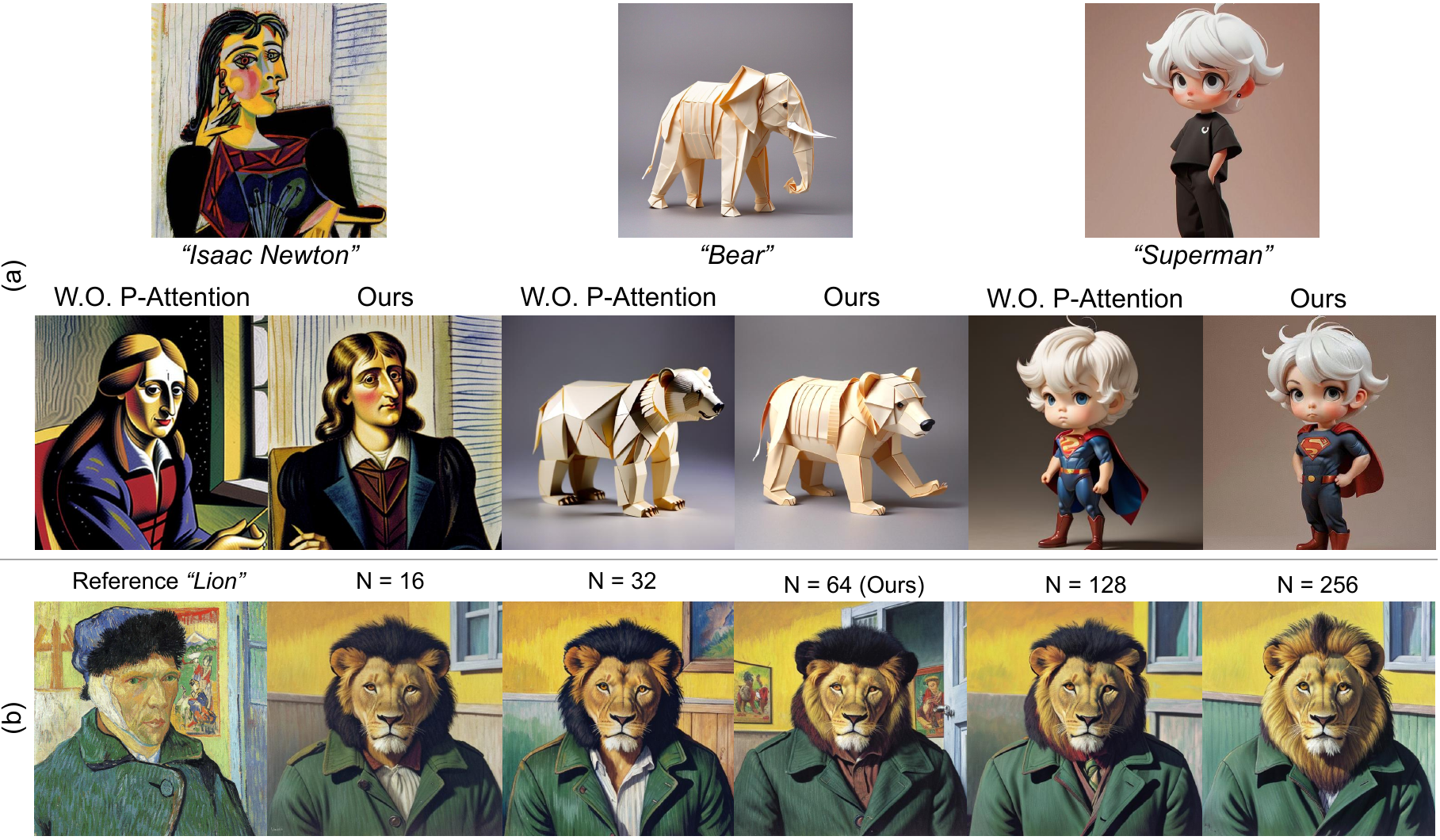}
\caption{
\textbf{Ablation studies for attention-based Style Extraction.} (a) Visual differences between \shortname generated samples without preceiver attention and our method. (b) Discussion on the number of learnable queries N.
}
\label{fig:resampler}
\end{figure}
\begin{figure}
\centering
\includegraphics[width= \linewidth]{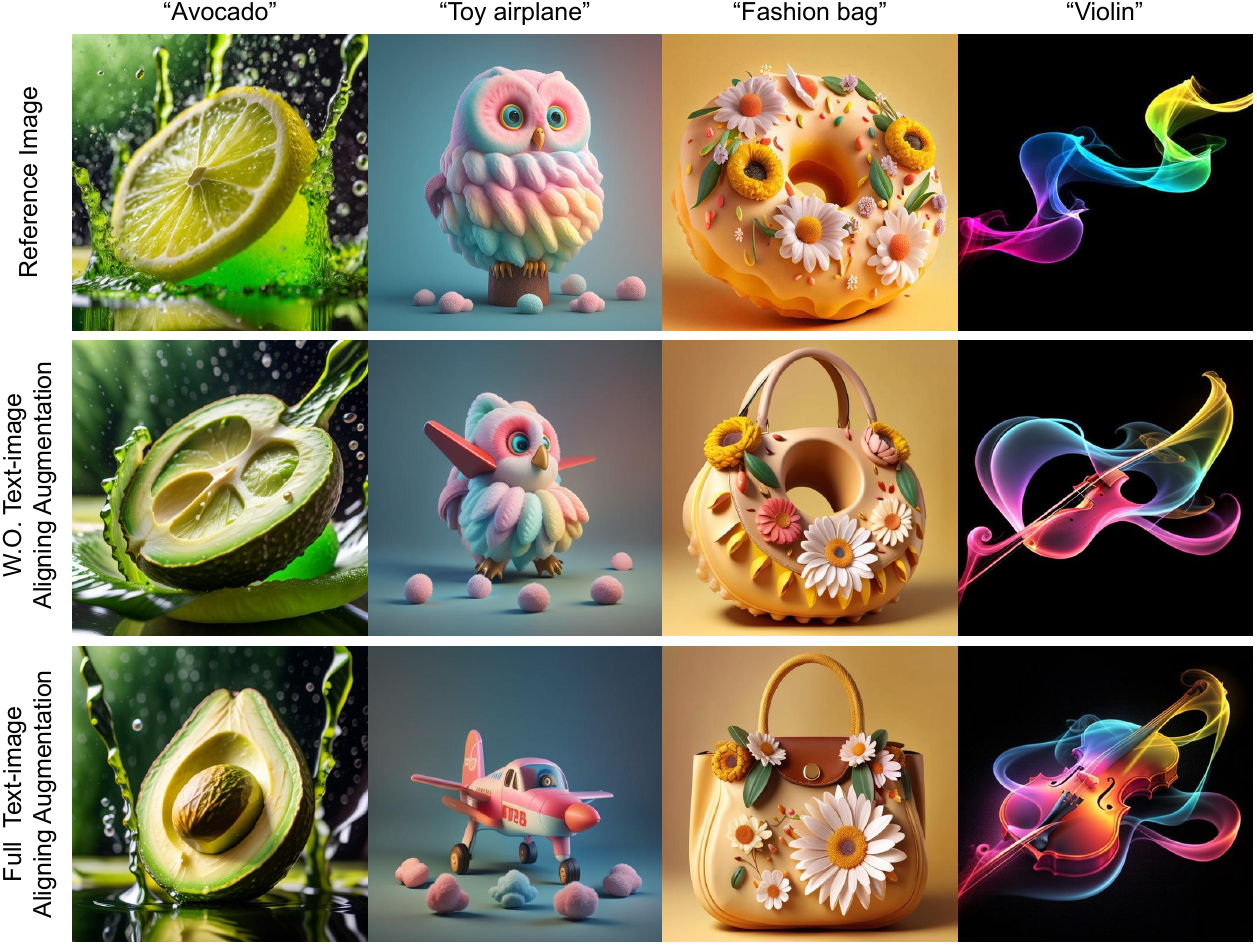}
\caption{\textbf{Ablation studies for text-image aligning augmentation.} All results use the same seed as well as setup factors. 
From the results at the bottom of the figure, it can be observed that this component plays a key role in the function of text conditioning.
}
\label{fig:attnenhanced}
\end{figure}
\begin{figure}
\centering
\includegraphics[width= \linewidth]{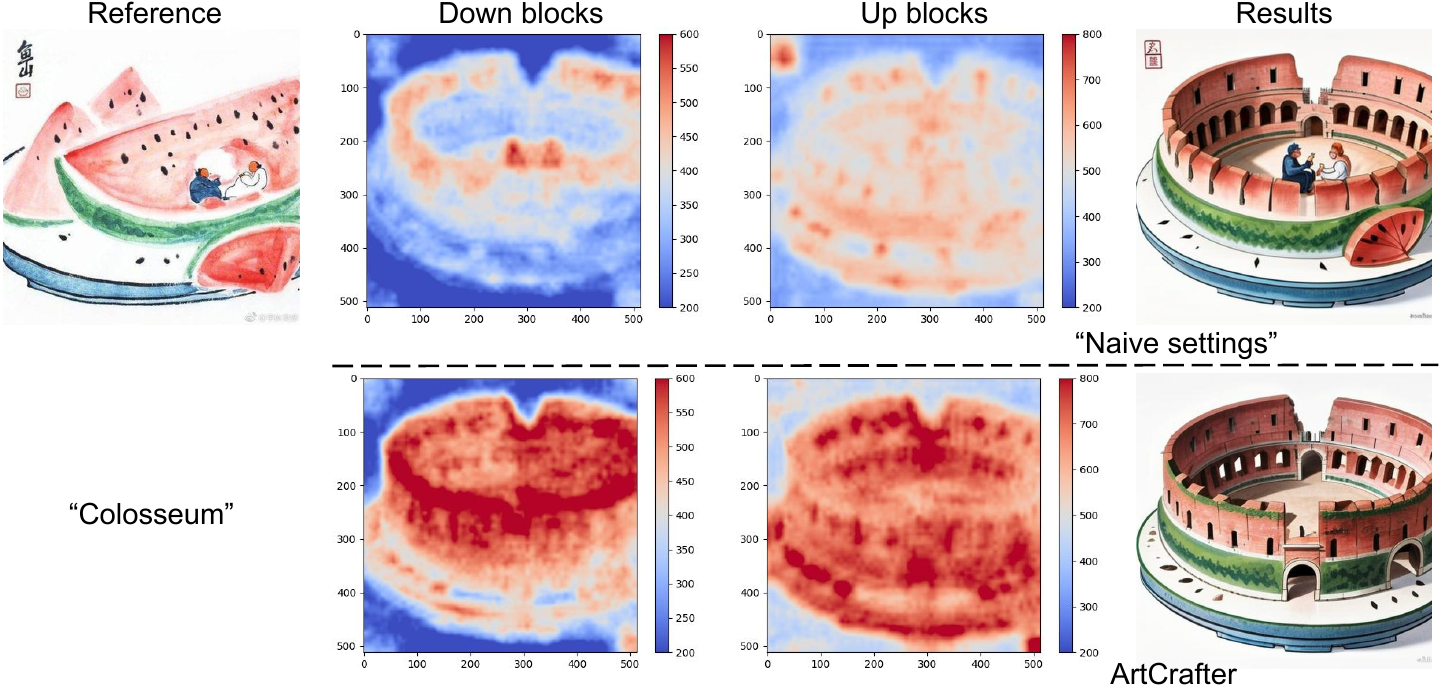}
\caption{\textbf{Ablation studies for text-image aligning augmentation.} Results of the “Naive setting” show a tendency to favor the reference image content over the textual content and fail to adequately reconstruct the target content. With the addition of the text-image aligning augmentation module (bottom), the rearranged attention enables more favorable text-based content reconstruction.
}
\label{fig:attn_visualization}
\end{figure}

\subsection{Quantitative Evaluations}
\noindent\textbf{Metric Description.}
To comprehensively assess the quality of the models, we computed the following metrics:
\textbf{CLIP-Text~\cite{clip}} measures the textual consistency of the generated image with the target description by calculating the cosine similarity between the target caption and the generated image.
\textbf{CLIP-Image~\cite{clip}} evaluates the similarity of the generated image to the target style by calculating the cosine similarity between the target styling image and the generated image.
\textbf{DINO-v2~\cite{dinov2}} further verifies the style consistency of the generated image with the target style by calculating the feature similarity between the target style image and the generated image.
\textbf{LPIPS~\cite{lpips}} measures the perceived similarity between two generated images, with higher values indicating less image similarity and better diversity.

As shown in Table~\ref{tab_quan_con}, our method outperformed other text-guided stylization methods in the CLIP-Text, CLIP-Image, DINO-v2, and LPIPS metrics. This indicates that our method has a significant advantage in generating images that are highly consistent with the target description and style. Moreover, our method effectively retains the details and semantics of the content while maintaining high stylization quality as well as diversity.

\subsection{User Study}
To obtain a more comprehensive assessment, we conducted a user study. We randomly selected $30$ generated results for each method covering a wide range of styles and hired $26$ professionals in the art field to evaluate these generated results.
Specifically, they rated text and image consistency on a scale of $1-5$, resulting in the Human-Text and Human-Image scores in Table ~\ref{tab_quan_userstudy}. The images with the best overall results are then pulled out to obtain the Human-Overall percentage results.
The results of Table~\ref{tab_quan_userstudy} indicate that the results generated by our proposed \shortname are more favored in all three aspects.
We notice the difference between objective evaluation metrics and subjective evaluation metrics because the former assesses each aspect in isolation.
Whereas users may integrate information across various aspects, despite separate options provided. Human Preference suggests that our generated results have struck a better balance among text consistency, image consistency, and overall visual appeal.

\subsection{Ablation Study}

\begin{figure}
\centering
\includegraphics[width= \linewidth]{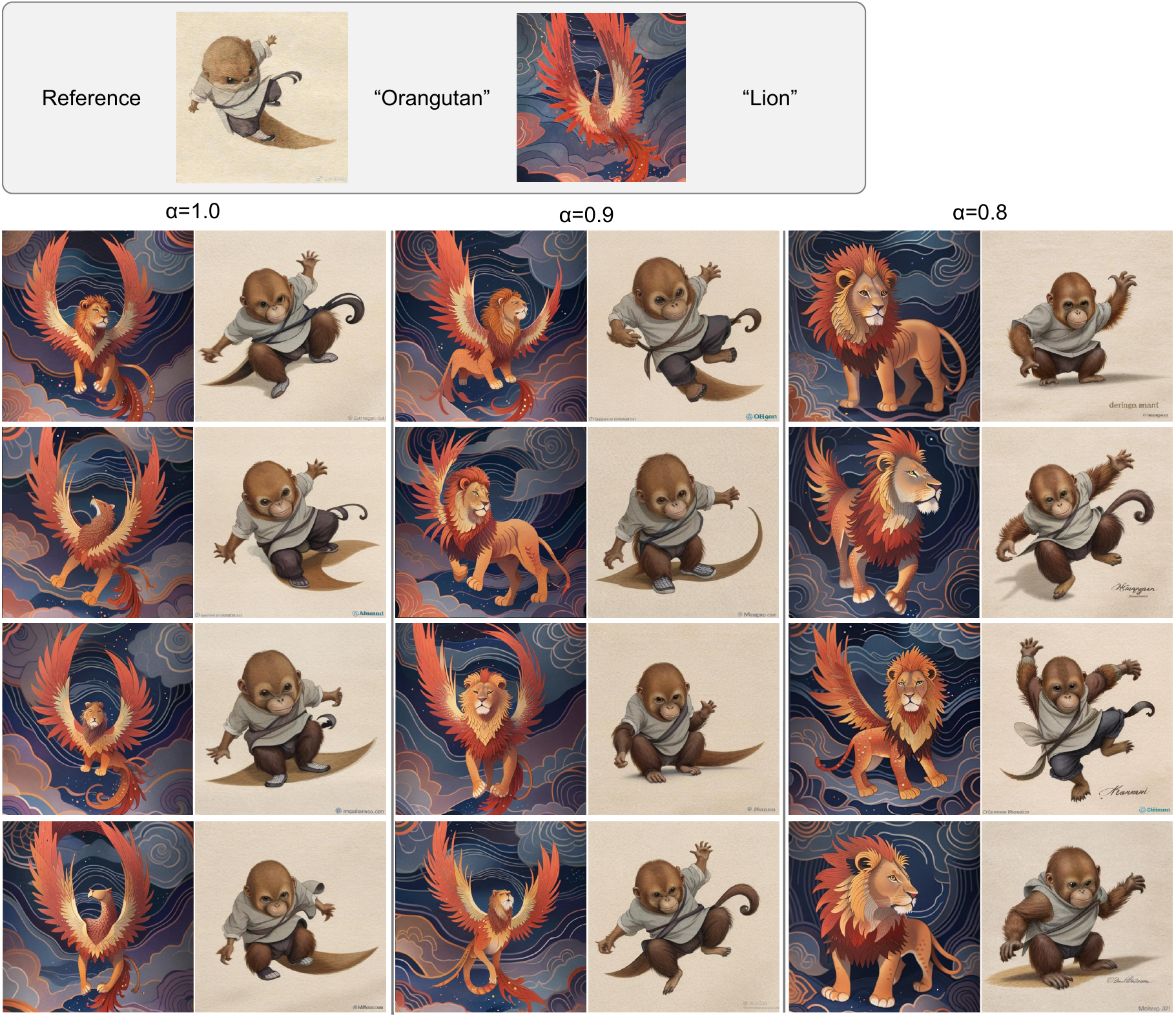}
\caption{\textbf{The effect of varying degrees of explicit modulation.} As $\alpha$ is increased, more varied results can be obtained (right columns), with some styles of misalignment of details.
}
\label{fig:linear}
\end{figure}
\begin{figure}
\centering
\includegraphics[width= \linewidth]{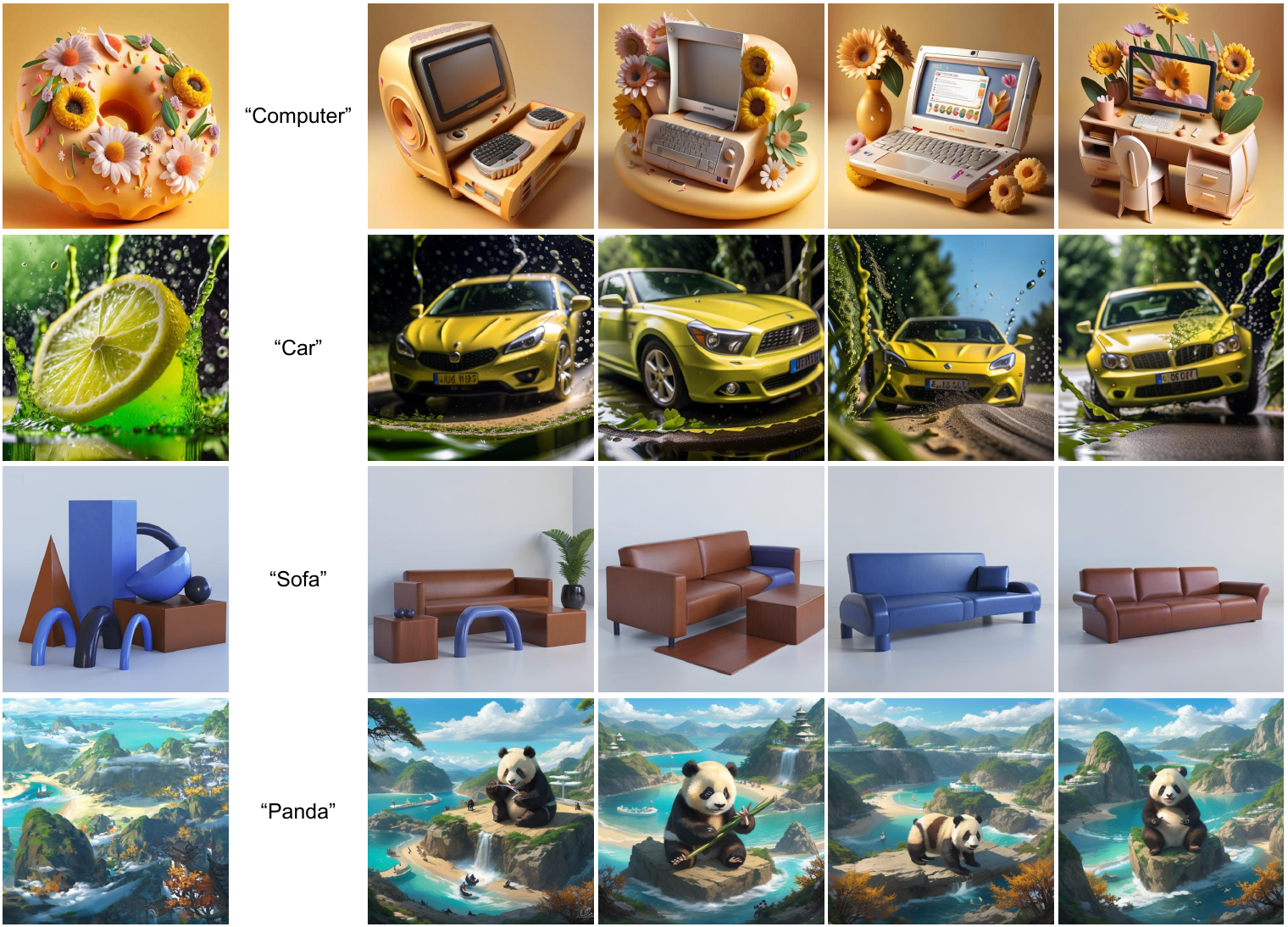}
\caption{\textbf{Diversity results with explicit modulation ablation study.} We conducted an ablation study to assess the impact of the explicit modulation component on the diversity of generated images. The results demonstrate that the explicit modulation component significantly enhances the diversity of the outputs.
}
\label{fig:sup_ablation_diversity}
\end{figure}

\subsubsection{Visualization analysis.}
\noindent\textbf{Attention-based style extraction.}
In our study, we observed that naive cross-attention tends to cause a shift in style, as shown in the results of Fig.~\ref{fig:compare} (Styleshot, Stylealigned, Stylecrafter, CSGO). The novel approach is by processing the embedded image's latent features with perceptual attention (Sec.~\ref{sec:3.2}). This method not only effectively prevents unintended style shifts but also significantly enhances artistic expression, particularly in terms of color and texture. As demonstrated in Fig.~\ref{fig:resampler} (a), the images processed with perceptual attention show a remarkable improvement in the richness of color and the detail of texture, making the generated images more artistically appealing and visually attractive.

To further optimize the performance of our model, we conducted an in-depth ablation study on the key parameter \( N \). As shown in Figure~\ref{fig:resampler} (b), we found that the value of \( N \) has a crucial impact on the model's attention distribution and generation results. When \( N \) is low, the model's attention becomes overly focused on narrow feature regions, leading to the loss of feature information and resulting in images that appear overly localized and lack a sense of coherence. Conversely, when \( N \) is high, the model's attention becomes too dispersed, blurring the key attributes of the noise and causing the generated images to lack clear structure and detail. Through this series of experiments, we identified an optimal balance point, enabling the model to capture and reproduce key image features accurately while maintaining the overall style.

\begin{table}
\centering
\footnotesize
\caption{\textbf{Quantitative ablation study of proposed components.} The results provide insights into how each component affects the performance of key metrics such as image consistency, text consistency, and diversity.}
\setlength{\tabcolsep}{4pt} 
\begin{tabular}{c|cccc}
\toprule
\textbf{Configuration} & \textbf{CLIP-Text} $\uparrow$ & \textbf{CLIP-Image} $\uparrow$ & \textbf{DINO-v2} $\uparrow$ & \textbf{LPIPS} $\uparrow$    \\
\midrule
W.O. ASE~\ref{sec:3.2} &23.49    &66.81  &37.25   &0.4971    \\
W.O. TIAA~\ref{sec:3.3}  &19.13   &70.52  &41.77   &0.4316 \\
W.O. EM~\ref{sec:3.4}  &22.19   &71.39  &42.63    &0.3572 \\
\bottomrule
\end{tabular}

\label{tab_ablation_study}
\end{table}

\noindent\textbf{Text-image aligning augmentation.}
To evaluate the effectiveness of the TIAA (Sec.~\ref{sec:3.3}), we present visual representations of the stylization results in Fig.~\ref{fig:attnenhanced} and Fig.~\ref{fig:attn_visualization}. Fig.~\ref{fig:attnenhanced} clearly illustrates that the TIAA component significantly enhances the alignment between text and image, thereby improving the overall efficacy of the text role. Moreover, the heatmap in Fig.~\ref{fig:attn_visualization} reveals the cosine similarity between the cross-attention of the generated results and the textual self-attention results. This indicates that ArtCrafter, through its fine-grained attention mechanism, effectively aligns the generated images with the textual descriptions. This visual evidence strongly supports the success of our strategy in achieving tighter compatibility and refinement between text and images via the TIAA module.

\begin{figure*}[ht]
\centering
\includegraphics[width= \linewidth]{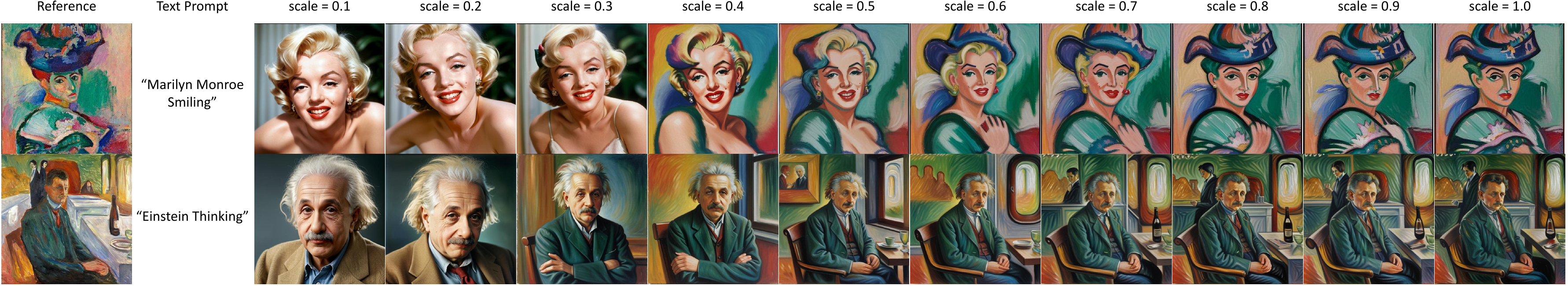}
\caption{\textbf{Visual demonstrations of \shortname's influence on pre-trained diffusion models.} The illustrations reveal how \shortname's modulation enhances the model's ability to generate images that are stylistically coherent with varying degrees of influence. As the intensity of \shortname's application increases, there is a noticeable evolution in the image's stylistic attributes, demonstrating the adaptability and controllability of our method.
}
\label{fig:scale}
\end{figure*}
\begin{figure}
\centering
\includegraphics[width= \linewidth]{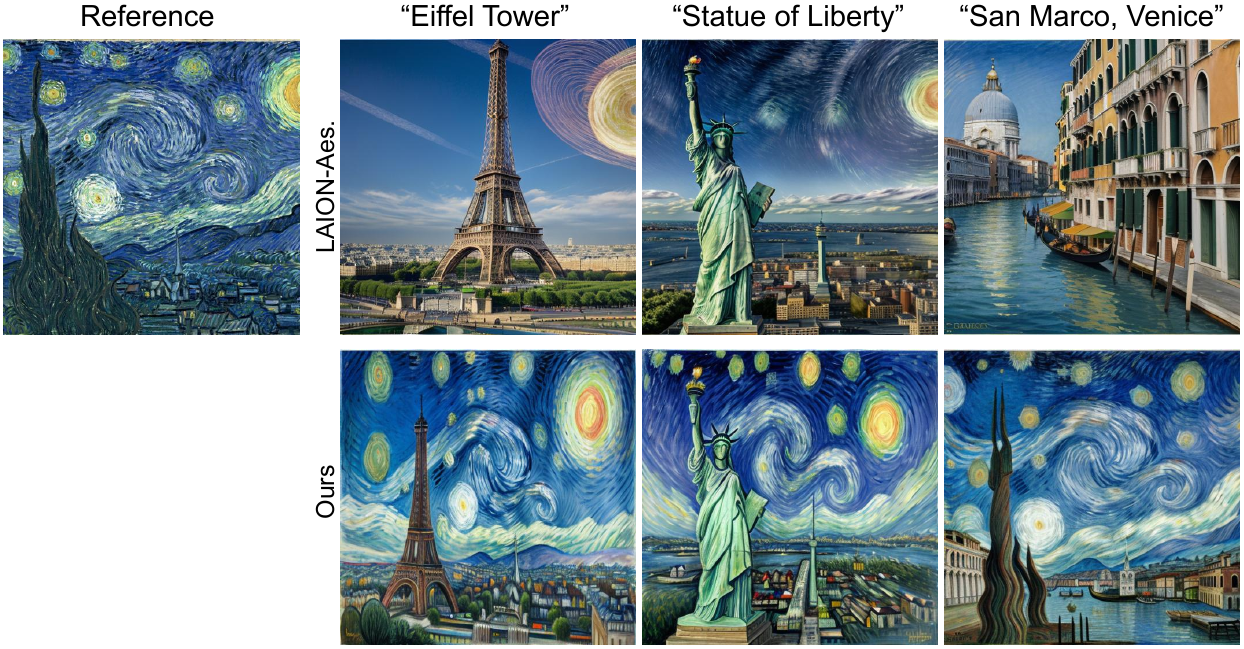}
\caption{\textbf{Visualization results of \shortname architecture trained on different datasets.} Specifically, when trained on a dataset with rich artistic styles, \shortname generates images with intricate and varied artistic features. In contrast, when trained on a dataset with more natural images, the model produces outputs that are highly consistent with natural visual elements. These results underscore the robustness and versatility of \shortname in generating high-quality images.}
\label{fig:dataset}
\end{figure}

\noindent\textbf{Explicit modulation.}
We investigate the impact of the dynamic scale \( \alpha \) in EM (Sec.~\ref{sec:3.4}) on the diversity of generated results by adjusting its value. As illustrated in Fig.~\ref{fig:linear}, increasing \( \alpha \) leads to a more diverse set of images while maintaining common stylistic attributes with the reference image. We typically set \( \alpha \) to \( 0.8 \).

Fig.~\ref{fig:sup_ablation_diversity} provides detailed ablation results for our EM component, further demonstrating its effectiveness in enhancing content diversity. These results highlight the component's ability to significantly enrich the variety of outputs, ensuring a broader spectrum of creative expressions. The experiments detailed in Fig.~\ref{fig:sup_ablation_diversity} underscore how EM contributes to generating images that not only capture the essence of the input text but also exhibit a wide range of artistic styles, thereby validating its importance in achieving diverse and compelling visual outcomes.

\subsubsection{quantitative analysis.}
To fully evaluate the performance of the proposed components, we conduct an ablation study using both quantitative and qualitative methods. 
Table~\ref{tab_ablation_study} summarizes the quantitative results, in which we test ArtCrafter's performance after removing each essential technique component individually.
There are inherent trade-offs between image consistency, text consistency, and diversity. Each module enhances a specific aspect: ASE improves image performance (CLIP-Image, DINO-v2), TAA boosts text performance (CLIP-Text), and EM parametrically balances and diversifies the results (LPIPS). Our goal is to achieve an optimal balance among these aspects to maximize overall effectiveness.
Collectively, these findings underscore the significance and efficacy of each component within the overall strategy.





\subsubsection{ArtCrafter control.}
We test the role of ArtCrafter in the pre-training diffusion model.
The result of adjusting the scale $\alpha$ of ArtCrafter in the pre-trained diffusion model is shown in Fig.~\ref{fig:scale}. 
Fig.~\ref{fig:scale} illustrates the outcomes at intervals of $0.1$ for the guidance coefficient, thereby providing a granular view of how varying levels of input influence the final generation quality. 
In this work, we usually set the scale to $0.6$.
This verifies that ArtCrafter can effectively enhance the applicability of pre-trained diffusion models in the art generation domain at a lower training cost. With this adjustment, we can more flexibly control the degree of text consistency and stylization of the generated images to meet different creative needs.

\subsubsection{Dataset.}
We constructed the \datasetname dataset using art images from WikiArt~\cite{wikiart} and LAION-Aesthetics~\cite{laion-aesthetics}, and utilized BLIP-2~\cite{blip2} as the textual style descriptive model to create art-description data pairs. To evaluate the impact of our dataset, we conducted an ablation study by training the model solely on the LAION-Aesthetics dataset. As shown in Fig.~\ref{fig:dataset}, the model trained only on LAION-Aesthetics tends to generate realistic landscape images. In contrast, our model trained on the combined dataset effectively captures the distinctive style of Van Gogh's \textit{Starry Night}. This demonstrates that incorporating diverse artistic data significantly enhances the model's ability to generate a wide range of artistic styles.

\begin{figure*}
\centering
\includegraphics[width=0.65\linewidth]{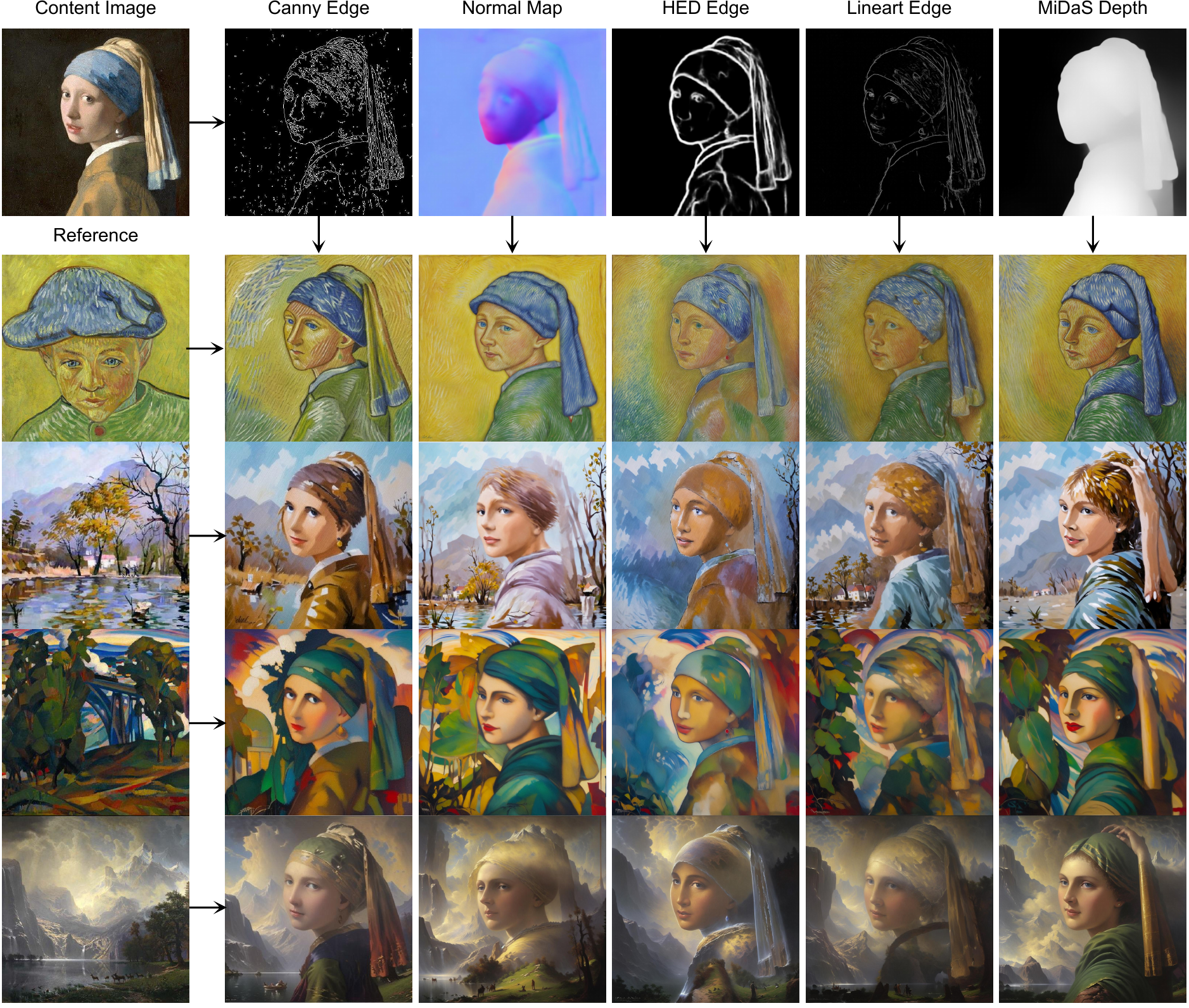}
\caption{\textbf{Integration of \shortname with additional conditions.} The outcomes of this integration indicate that \shortname not only effectively incorporates images as content guidance for style transfer but also seamlessly integrates with 3D content information. This compatibility showcases \shortname's potential to enhance the richness of generated images by leveraging various types of input data, thereby expanding its applicability in diverse creative scenarios.
}
\label{fig:controlnet}
\end{figure*}

\section{Discussion}
\subsection{Application}
\subsubsection{ArtCrafter with additional conditions applied.}
Benefiting from our design without any changes to the network structure of the original diffusion model, ArtCrafter is seamlessly compatible with existing controllable tools~\cite{controlnet}. Fig.~\ref{fig:controlnet} shows the diverse examples generated by applying different structural controls, including canny edge detection~\cite{canny}, normal map~\cite{normal}, HED edge detection~\cite{hed}, lineart edge, and MiDaS depth map~\cite{midas}. These examples not only highlight ArtCrafter's flexibility in adapting to a variety of conditions but also foretell its great potential for application in the $3D$ field, opening up new possibilities for artistic creation and visual design.

\begin{figure*}[ht]
\centering
\includegraphics[width= 0.85\linewidth]{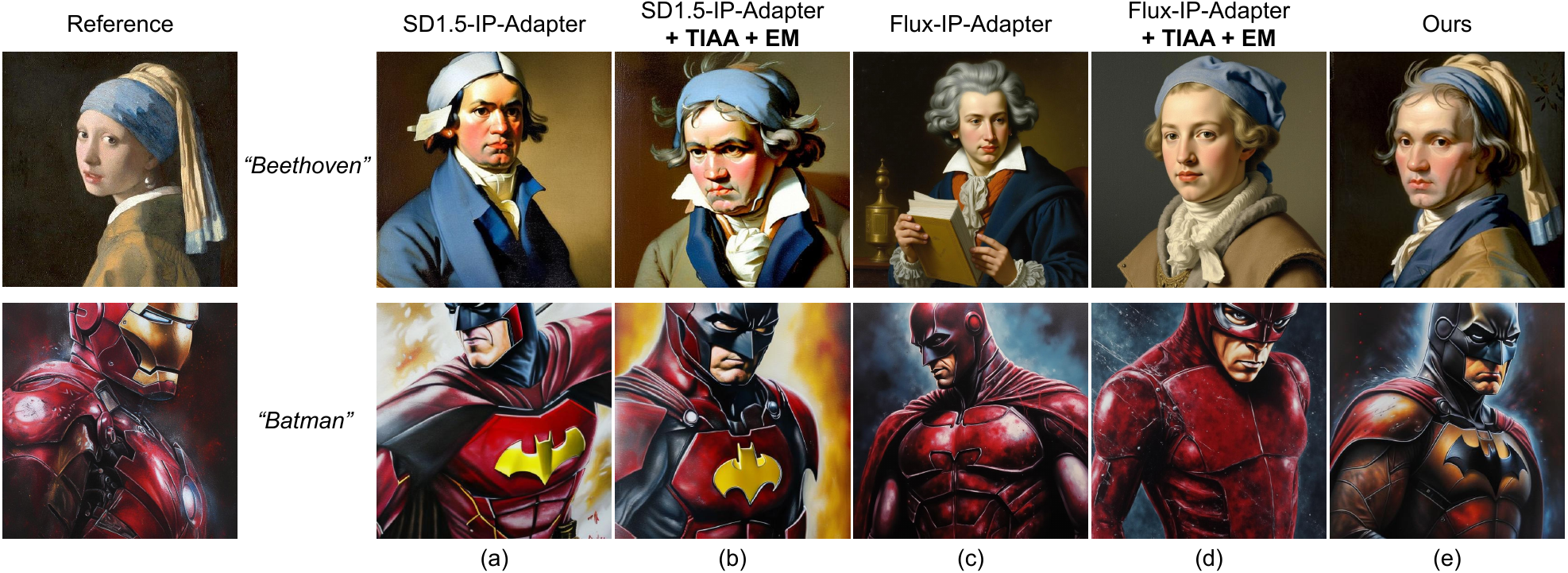}
\caption{\textbf{Enhanced comparisons with IP-Adapter variants.} To further validate the superiority of our approach, we conducted additional comparative analyses with the SD1.5 and Flux versions of the IP-Adapters, which were enhanced with our TIAA and EM algorithms. These comparisons provide a comprehensive view of how our algorithms contribute to the performance of these models.}
\label{fig:flux}
\end{figure*}
\subsubsection{More applications on other T2I models.}
%
%
%
To further verify the transferability and universality of the proposed method in multi-functional application scenarios, we conducted a series of in-depth comparative experiments on various text-to-image generation models with different weight configurations and architectural designs. As shown in Fig.~\ref{fig:flux}, we present the comparative results of two representative model versions—Stable Diffusion 1.5 (SD1.5) and Flux—after integrating two key modules of our method: the TIAA module and the EM module. The experimental results demonstrate that by embedding these two modules into the models in a lightweight integration manner, significant performance improvements can be achieved without the need for additional training or fine-tuning. This fully proves the transferability and efficiency of our method.

\begin{figure*}
\centering
\includegraphics[width= 0.9\linewidth]{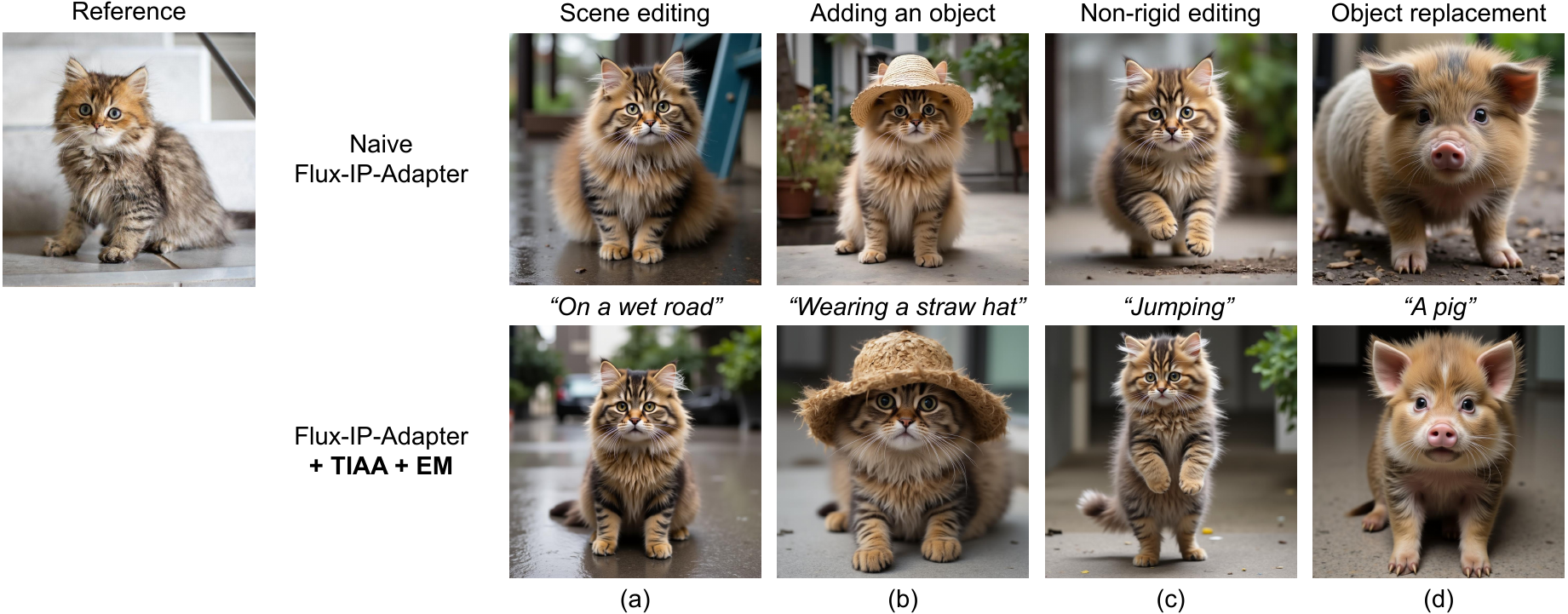}
\caption{\textbf{Versatility of our algorithms in natural image editing.} Our algorithms are designed to be versatile and applicable across a broad spectrum of natural image editing applications. They are effective regardless of the artistic dataset used, demonstrating their robustness and adaptability. This capability allows for a seamless integration into various workflows, enhancing the potential for creative expression.}
\label{fig:application_editing}
\end{figure*}

\subsubsection{Natural image editing.}
The strengths of ArtCrafter are not confined to its training on the ArtMarket dataset; its performance in various natural image editing scenarios is also highly impressive. Fig.~\ref{fig:application_editing} demonstrates the remarkable capabilities of TIAA and EM in jointly enhancing image-text guidance. Our method can be widely applied to multiple natural image editing scenarios, including scene editing (such as adjusting the atmosphere or weather of a scene), adding objects (naturally integrating new objects into an image), non-rigid editing (deforming objects), and object replacement (replacing one object in an image with another).
In Fig.~\ref{fig:application_editing} (a), the improved results more accurately reflect the "road" information from the text, such as more clearly presenting the texture, direction, or surrounding environment of the road. In Fig.~\ref{fig:application_editing} (b), (c), and (d), not only are the requirements of the editing text better enhanced, but the information of non-edited parts, such as the details of the indoor environment, is also preserved, ensuring the naturalness and coherence of the overall image. This capability makes ArtCrafter valuable and broadly applicable in the field of natural image editing.

\begin{figure}
\centering
\includegraphics[width= \linewidth]{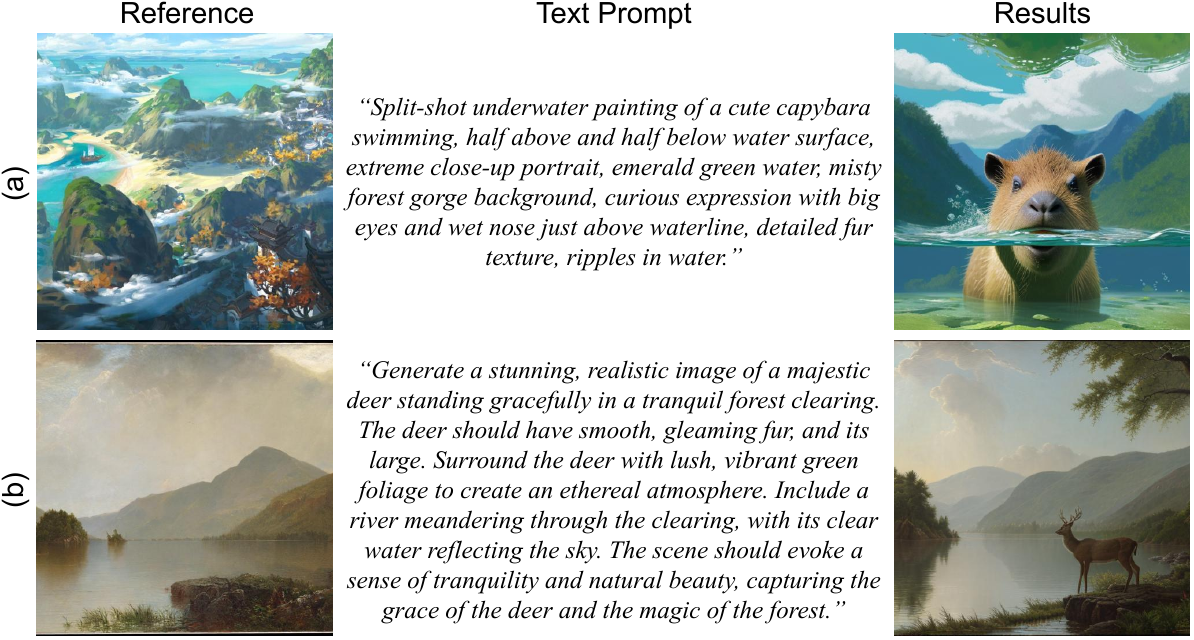}
\caption{\textbf{Enhanced comprehension of extended text.} Our model demonstrates an advanced ability to interpret and understand lengthy and intricate textual descriptions. }
\label{fig:longtext}
\end{figure}
\begin{figure}
\centering
\includegraphics[width= \linewidth]{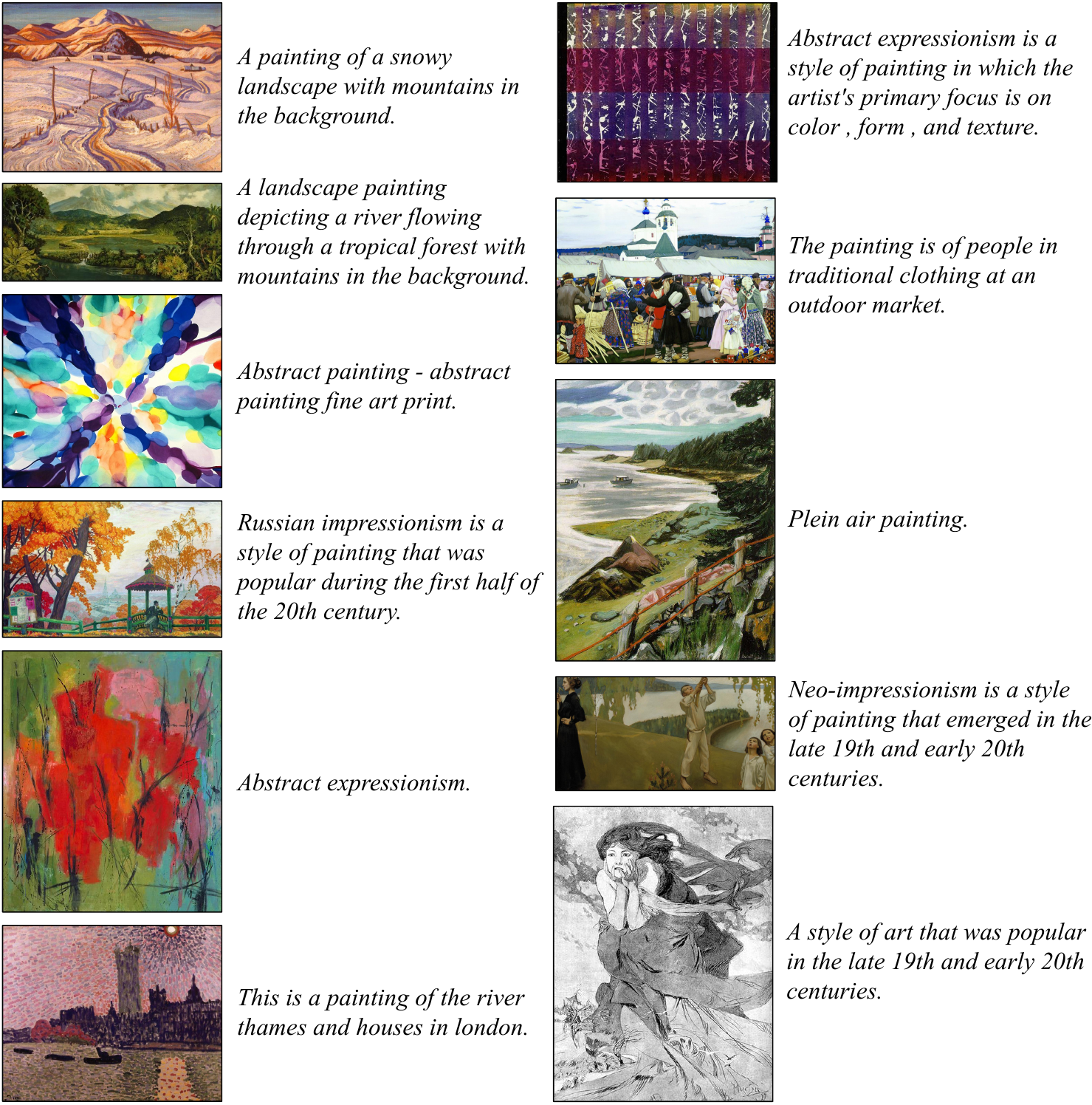}
\caption{\textbf{Dataset samples.} The dataset samples cover a diverse range of artistic styles and periods, ensuring a comprehensive representation of the artistic spectrum. The textual descriptions are crafted to capture the essence of each artwork, providing detailed insights into the visual elements, thematic content, and stylistic nuances.}
\label{fig:dataset}
\end{figure}

\subsubsection{Long Text Prompt}
%
%
In Fig.~\ref{fig:longtext}, we present a series of long and complex textual requirements that our model successfully translates into detailed visual outputs. For instance, in Fig.~\ref{fig:longtext} (a), the model accurately generates an image with a deer having a ``curious expression with big eyes and a wet nose just above the waterline'', ``detailed fur texture'', ``and ripples in the water'', capturing the ``half above and half below water surface'' scenario. In Fig.~\ref{fig:longtext} (b), the model creates a serene forest scene with a deer ``surrounded by lush foliage,'' ``a river meandering,'' and ``clear water reflecting the sky.'' These examples demonstrate the robust capability of our text-image alignment design in handling intricate textual instructions and generating images that precisely match the detailed descriptions.

\begin{figure*}[ht]
\centering
\includegraphics[width= 0.9\linewidth]{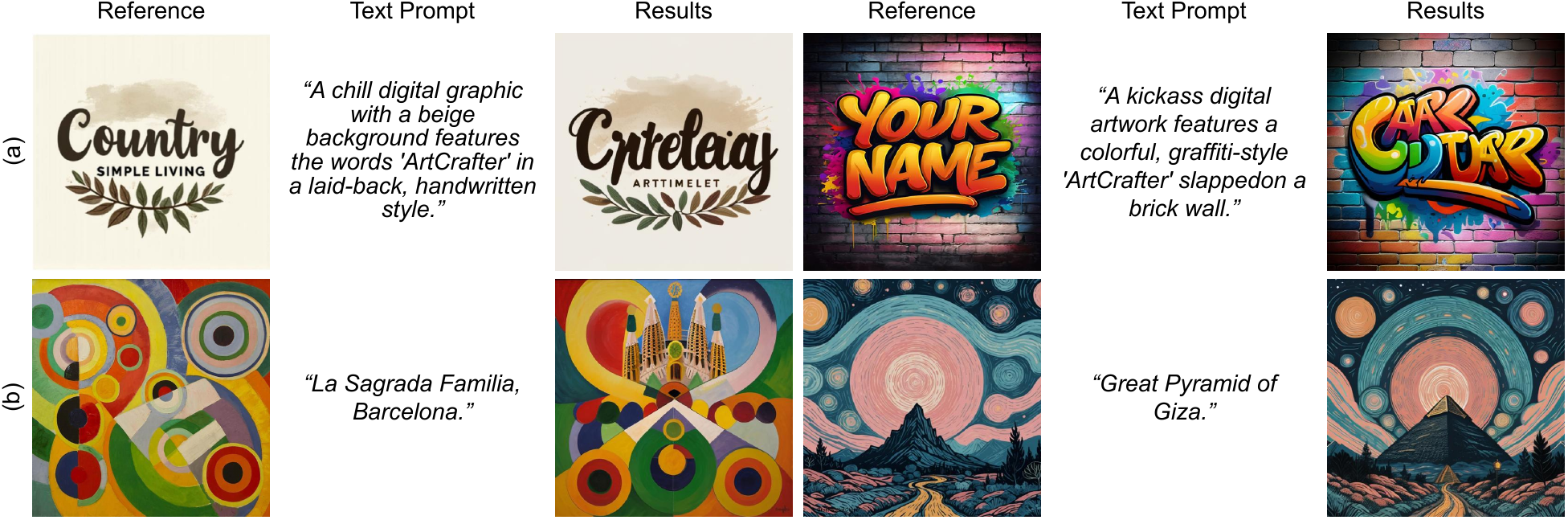}
\caption{\textbf{Analysis of failure cases.}  (a) Text generation failed. (b) Instances of inconsistent line texture.}
\label{fig:failurecases}
\end{figure*}

\subsection{Dataset Samples}
Our dataset \datasetname, as shown in Fig.~\ref{fig:dataset}, contains a rich collection of artworks along with their corresponding textual descriptions. The visual content of the dataset is primarily sourced from the extensive art repositories of WikiArt~\cite{wikiart} and LAION-Aesthetics~\cite{laion-aesthetics}, which cover a wide range of artistic styles and periods. The descriptive texts are derived from BLIP-2~\cite{blip2}, an advanced language model that can generate detailed and contextually rich captions. The combination of high-quality images and accurately descriptive texts enables \datasetname to effectively support the training of models for style transfer and content generation in the field of text-to-image synthesis.

\subsection{Limitation}
%
%
%
%
The limitations of ArtCrafter are mainly reflected in two aspects. First, it is the ability to accurately generate text. As shown in Fig.~\ref{fig:failurecases} (a), ArtCrafter has difficulty in accurately generating text in images and cannot ensure the clarity and accuracy of the text content ``ArtCrafter''. This limitation may seriously affect its performance in image generation tasks that require text elements. For example, in designing posters, brochures, or other image editing tasks that require precise text layout, ArtCrafter may not be able to meet the high-precision requirements of users for text content and layout.
Secondly, the model can only partially follow large-scale textures and brushstrokes. For example, in Fig.~\ref{fig:failurecases} (b), our method incorrectly transforms the large-scale curves from left to right in the image into inconsistent curve shapes. This error indicates that the model may fail to fully understand and accurately reproduce the original shapes and directions when dealing with complex textures and brushstrokes. Accurately replicating complex long-range texture patterns is one of the challenges for our model to fully assimilate and reproduce in the generated output.

\section{Conclusions and Future Work}
In this paper, we introduce \shortname, a novel text-image aligning style transfer framework achieved through an embedding reframing architecture. Our approach ensures superior text-guided style transfer quality by integrating three core components: attention-based style extraction, text-image aligning augmentation, and explicit modulation. Comprehensive evaluations demonstrate \shortname strengths in adapting to diverse artistic styles, maintaining textual prompt consistency, enhancing output diversity, and improving overall visual quality.

Acknowledging the current limitations of our work, for future research, we intend to enhance our approach by incorporating pattern reproducibility and contextual elements within style images, including the relative positioning of style patches, to facilitate a more cohesive art style transfer. We expect that advancements in the extraction and fusion of style and content features, coupled with an investigation into the method's scalability and adaptability, will markedly enhance the quality of style transfer and provide more precise control over the shape and appearance similarity of the generated images.

\bibliographystyle{IEEEtran}
\bibliography{references}


\end{document}